\documentclass[11pt]{article}

\usepackage[final]{acl}

\usepackage{times}
\usepackage{latexsym}

\usepackage[T1]{fontenc}
\usepackage[utf8]{inputenc}

\usepackage{microtype}

\usepackage{inconsolata}

\usepackage{graphicx}
\usepackage{amssymb}
\usepackage{amsmath}
\usepackage{svg}
\usepackage{tabularx}
\usepackage{booktabs}
\usepackage{multirow}
\usepackage{enumitem}

\usepackage{hyperref}

\title{Measuring What Matters Beyond Text: Evaluating Multimodal Summaries by Quality, Alignment, and Diversity}

\author{
  Abid Ali\thanks{Corresponding Author} \and
  Diego Moll{\'a}-Aliod \and
  Usman Naseem \\
  School of Computing, Macquarie University, Sydney, Australia \\
  \texttt{abidmeeraj@gmail.com, diego.molla-aliod@mq.edu.au, usman.naseem@mq.edu.au}
}

\begin{document}
\maketitle

\begin{abstract}

Multimodal Large Language Models (MLLMs) have facilitated Multimodal Summarization with Multimodal Output (MSMO), wherein systems generate concise textual summaries accompanied by salient visuals from multimodal sources. However, current MSMO evaluation remains fragmented: text quality, image--text alignment, and visual diversity are typically assessed in isolation using unimodal metrics, making it difficult to capture whether the modalities jointly support a faithful and useful summary. To address this gap, we introduce \texttt{MM-Eval}, a unified evaluation framework that integrates assessments of textual quality, cross-modal alignment, and visual diversity. \texttt{MM-Eval} comprises three components: (1) text quality, measured using OpenFActScore for factual consistency and G-Eval for coherence, fluency, and relevance; (2) image--text relevance, evaluated via an MLLM-as-a-judge approach; and (3) image-set diversity, quantified using Truncated CLIP Entropy. We calibrate \texttt{MM-Eval} through a learned aggregation model trained on the \textit{mLLM-EVAL} news benchmark, aligning component contributions with human preferences. Our analysis reveals a text-dominant hierarchy in this setting, where factual consistency acts as a critical determinant of perceived overall quality, while visual relevance and diversity provide complementary signals. \texttt{MM-Eval} improves over heuristic aggregation baselines and provides an interpretable, reference-weak framework for comparative evaluation of multimodal summaries.

\end{abstract}

\section{Introduction}
Modern information consumption increasingly combines text with visuals, motivating Multimodal Summarization with Multimodal Output (MSMO): the task of generating textual summaries paired with relevant images that enrich and contextualize the content \cite{Zhu2018MSMO:Output, Zhu2020MultimodalReference}. MSMO aims to produce coherent multimodal outputs that balance informativeness with cognitive load \cite{Jiang2023ExploitingSummarization}, improving user satisfaction through faster gist comprehension via images while preserving detail in text \cite{Zhu2018MSMO:Output}.

Despite its promise, MSMO has historically been constrained by the semantic gap: the challenge of aligning low-level visual signals with high-level textual semantics. Recent advances in Multimodal Large Language Models (MLLMs), such as GPT-4V, LLaVA \cite{Liu2023VisualTuning}, and Qwen-VL \cite{Wang2024Qwen2-VL:Resolution}, have helped bridge this gap by supporting joint encoding and reasoning across modalities \cite{Chang2024AModels}. However, as generative capabilities improve, a critical bottleneck has emerged: the lack of robust, scalable, and reliable automatic evaluation methods for multimodal outputs \cite{Zhuang2024AutomaticOutput}.

Despite the inherently multimodal nature of MSMO, current evaluation protocols remain fragmented: a limitation we refer to as the ``Silo Effect''. Outputs are decomposed and assessed via unimodal metrics, each blind to cross-modal coherence. On the textual side, ROUGE remains dominant. While historically impactful, ROUGE is limited in its ability to capture semantic equivalence or factual consistency \cite{Schluter2017TheRouge, Zhang2024Fine-grainedSummarization}. Consequently, summaries that superficially match reference texts but contain hallucinated facts can still score highly \cite{Lage2025Openfactscore:Generation}.

On the visual side, Image Precision (IP) evaluates whether the model selects the exact images chosen by annotators, penalizing semantically valid alternatives, and ignoring alignment with textual content (see Section~\ref{sec:lit_review} for a detailed discussion).

\begin{figure*}[!t]
\vspace{-0.2cm}
    \centering
    \includegraphics[width=0.85\textwidth]{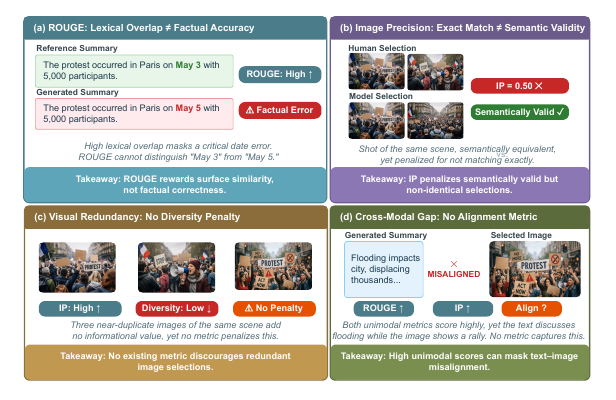}
    \vspace{-0.3cm}
    \caption{Toy examples showing limitations of current MSMO evaluation.}
    \label{fig:toy_example}
    \vspace{-0.4cm}
\end{figure*}

Most critically, current metrics fail to assess the interaction between modalities: a system may score well on ROUGE and image selection independently while producing disconnected multimodal outputs. Figure~\ref{fig:toy_example} illustrates these limitations with toy examples.

To address this, we introduce \texttt{MM-Eval}, a modular and unified evaluation framework designed to assess multimodal outputs along three key dimensions: quality, alignment, and diversity. Rather than combining existing tools, \texttt{MM-Eval} redefines evaluation criteria to reflect the inherently cross-modal nature of MSMO.

The framework consists of three pillars:
\begin{itemize}[noitemsep,leftmargin=*]
    \item \textbf{Text Quality} ($S_{text}$): We adopt a ``decompose-then-verify'' approach using OpenFActScore \cite{Lage2025Openfactscore:Generation} to assess atomic factual consistency, and G-Eval \cite{Liu2023G-EVAL:Alignment}, an LLM-based protocol for coherence, fluency, and relevance.

    \item \textbf{Cross-Modal Alignment} ($S_{relevance}$): Using an MLLM-as-a-Judge framework \cite{Zhuang2024AutomaticOutput}, we measure how well selected images semantically support the generated text, capturing the integrated quality of the multimodal summary.

    \item \textbf{Visual Diversity} ($S_{diversity}$): To discourage visual redundancy, we use Truncated CLIP Entropy (TCE) \cite{Ibarrola2024MeasuringGeneration}, which measures the distinctiveness of the image set in latent space.
    
\end{itemize}

\texttt{MM-Eval} aggregates these components via a learned weighting model trained on the \textit{mLLM-EVAL} news benchmark. Our analysis of human judgments from this dataset reveals a text-dominant preference structure in the news domain. In news, factual consistency serves as a key determinant of perceived quality, while visual relevance and diversity provide complementary signals whose contribution is conditional on adequate textual fidelity. We validate these findings through a human evaluation study that confirms annotators value all three dimensions, supporting MM-Eval's multi-pillar design. Since the component scorers are reference-weak and domain-portable, \texttt{MM-Eval} can be applied to new domains by recalibrating only the lightweight aggregation weights with minimal human supervision. This paper details the design of \texttt{MM-Eval}, its component metrics, and the empirical patterns uncovered through alignment with human ratings.

\section{Literature Review}\label{sec:lit_review}

\subsection{Traditional Evaluation Metrics}
\paragraph{Text-based Metrics.}

Summarization evaluation has long relied on ROUGE \cite{Lin2004ROUGE:Summaries}, which computes $n$-gram overlap with reference summaries. While effective for extractive systems, ROUGE cannot capture semantic equivalence, paraphrasing, or factual errors, and penalizes bridging sentences that refer to accompanying visuals \cite{Schluter2017TheRouge}. BERTScore \cite{Zhang2019BERTScore:Bert} addresses lexical rigidity via contextual embeddings but remains reference-based and fails to distinguish factual from hallucinated content \cite{Zhang2024Fine-grainedSummarization, Wan2022EvaluatingSummarization}.

\paragraph{Image-based Metrics.}
Image selection is typically evaluated using Image Precision (IP) and Image Recall (IR), which check whether the model selects the exact annotator-chosen images \cite{Zhu2020MultimodalReference}. This assumes a single correct reference set: if a model selects a semantically equivalent but different image, it receives no credit. Moreover, redundant images in the reference set may inflate IP without adding informational value.

\paragraph{Multimodal Aggregation.}
The Multimodal Automatic Evaluation (MMAE) framework \cite{Zhu2018MSMO:Output} aggregates ROUGE-L, IP, and image--text cosine similarity via a regression model trained on human satisfaction scores. While the aggregation principle is sound, reliance on ROUGE and IP retains the shortcomings mentioned above. \texttt{MM-Eval} builds on this idea but incorporates semantically grounded, reference-weak components.

\subsection{LLM-based Evaluation Approaches}

\paragraph{LLM-as-a-Judge.}
Recent work such as \textit{mLLM-Eval} \cite{Zhuang2024AutomaticOutput} uses large multimodal models (e.g., GPT-4V) to rate summaries directly via Chain-of-Thought prompting.
While effective, this approach is resource-intensive and acts as a black box with no explicit quality breakdown. \texttt{MM-Eval} addresses these gaps by decomposing evaluation into interpretable, modular sub-metrics built on open-source models, allowing individual components to be independently replaced or upgraded.

\paragraph{Fact-level and Diversity Evaluation.}
FActScore \cite{Min2023Factscore:Generation} proposes a ``decompose-then-verify'' approach, breaking summaries into atomic facts validated against the source. OpenFActScore \cite{Lage2025Openfactscore:Generation} extends this with open-source models. For visual diversity, Truncated CLIP Entropy (TCE) \cite{Ibarrola2024MeasuringGeneration} computes entropy over eigenvalues of CLIP embeddings, providing a reference-free signal that penalizes redundant visual outputs without requiring large sample sets like FID \cite{Heusel2017GansEquilibrium}.
The extent to which spectral entropy in CLIP space fully captures human notions of visual diversity remains an open question.

\section{Methodology}

Formally, the input to \texttt{MM-Eval} consists of a source document $D = \{T_{\text{source}}, V_{\text{source}}\}$ and a system-generated summary $S_{\text{cand}} = \{T_{\text{gen}}, V_{\text{sel}}\}$. The output is a scalar score $S_{\text{final}} \in \mathbb{R}$, representing the overall quality of the candidate summary.

The score is computed by integrating signals from three core components: (1) textual quality, (2) cross-modal alignment, and (3) visual diversity.

\begin{figure*}[!t]
\vspace{-0.2cm}
    \centering
    \includegraphics[width=1\textwidth]{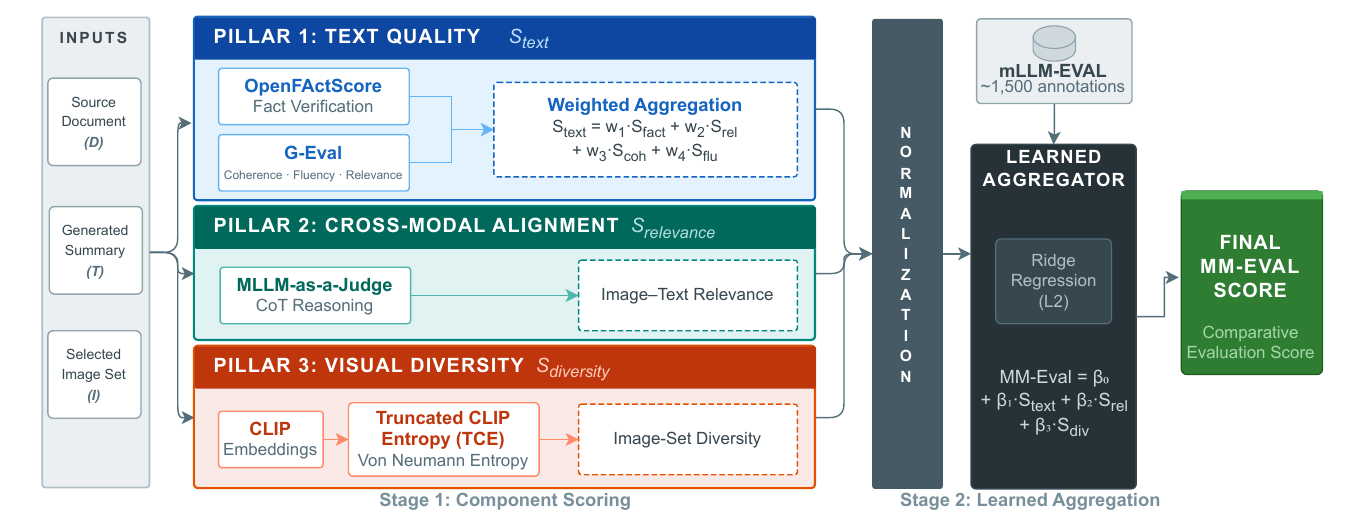}
    \vspace{-0.35cm}
    \caption{An overview of MM-Eval Framework.}
    \label{fig:framework}
    \vspace{-0.4cm}
\end{figure*}

\subsection{Pillar 1: Quantifying Textual Summary Quality ($S_{text}$)}
The textual dimension of \texttt{MM-Eval} assesses two core aspects: \textbf{Factual Consistency}: whether the generated statements are verifiable with the source; and \textbf{Qualitative Attributes} such as coherence, relevance, and fluency.

\subsubsection{Factual Consistency}
To identify hallucinations and ensure faithfulness to the source, we adopt the OpenFActScore pipeline, which shifts evaluation from $n$-gram recall to fact-level precision.

\paragraph{Step 1: Atomic Fact Generation (AFG).} The generated summary $T_{\text{gen}}$ is decomposed into a set of atomic facts $A = \{a_1, a_2, \dots, a_m\}$, where each $a_i$ expresses a single verifiable claim (e.g., ``The event occurred on Tuesday'') \cite{Lage2025Openfactscore:Generation}. We employ an instruction-tuned open-source LLM for this step.

\paragraph{Step 2: Atomic Fact Validation (AFV).} Each atomic fact $a_i$ is then validated against the source text $T_{\text{source}}$ using a second LLM, which functions as a binary classifier.

\paragraph{Step 3: Factuality Scoring.} The factual consistency score is computed as the average precision over validated facts:
\begin{equation}
    S_{fact} = \frac{1}{|A|} \sum_{i=1}^{|A|} v_i
\end{equation}

where $v_i \in \{0,1\}$ indicates whether $a_i$ is supported by the source. The resulting score ($0 \le S_{\text{fact}} \le 1$) is robust to paraphrasing and insensitive to output length \cite{Min2023Factscore:Generation}.

\subsubsection{Qualitative Metrics}
Subjective qualities such as relevance ($S_{\text{rel}}$), coherence ($S_{\text{coh}}$), and fluency ($S_{\text{flu}}$) are difficult to quantify using explicit formulas. We build on G-Eval \cite{Liu2023G-EVAL:Alignment}, an LLM-as-a-Judge approach shown to correlate strongly with human judgments. Standard LLM prompting can exhibit high variance (e.g., oscillating between adjacent scores), which G-Eval addresses through a structured form-filling protocol with Chain-of-Thought (CoT) and probability-weighted scoring.

Finally, $S_{text}$ is a weighted aggregation of all four intra-pillar components as below:
\begin{equation}\label{eq:s_text}
    S_{text} = w_1 S_{fact} + w_2 S_{rel} + w_3 S_{coh} + w_4 S_{flu}
\end{equation}

\subsection{Pillar 2: Assessing Image-to-Text Relevance ($S_{relevance}$)}
This pillar measures the alignment of the modalities. It answers the question: Do the selected images actually complement/supplement the text?

\subsubsection{MLLM-as-a-Judge for Alignment}
\texttt{MM-Eval}'s MLLM-as-a-Judge methodology captures the complex pragmatic relevance required by human annotators. Recent advancements show that MLLMs are capable of higher correlation with human relevance judgments than standard proxies. The MLLMs can be instructed to perform Chain-of-Thought (CoT) reasoning before assigning a score, mitigating the high variance observed in human preference \cite{Sahili2025FairJudge:Alignment}.

By stabilizing the measurement of cross-modal alignment, this component ensures the metric robustly evaluates whether the images truly add value to the summarized facts. Since \texttt{MM-Eval} is modular by design, the specific MLLM used for alignment scoring can be replaced as stronger open-source or proprietary models become available, without altering the rest of the framework.

\subsection{Pillar 3: Measuring Visual Diversity ($S_{diversity}$)}
A common failure mode in summarization is visual redundancy. For example, a news article about a protest might include multiple images captured from slightly different angles or taken seconds apart, each visually similar but not meaningfully distinct. A model might select several such near-duplicate images, resulting in low informational diversity. \texttt{MM-Eval} explicitly penalizes this behavior using spectral entropy encouraging more diverse and informative visual outputs.

\subsubsection{Truncated CLIP Entropy (TCE)}
Pillar 3 addresses the common multimodal failure mode of visual redundancy, and employs TCE to quantify $S_{diversity}$ using information theory.

The calculation of TCE involves generating CLIP embeddings ($F$) for the selected images and computing their empirical covariance matrix ($C$). The eigenvalues ($\lambda_i$) of $C$ represent the distribution of variance of the selected images across the principal axes of the semantic feature space \cite{Ibarrola2024MeasuringGeneration}. By normalizing these eigenvalues into a probability distribution ($p_i$) and then calculating the Von Neumann entropy
\begin{equation}
    S_{diversity} = - \sum_{i=1}^{k} p_i \log(p_i),
\end{equation}

the metric effectively measures the semantic volume spanned by the selected images in the CLIP embedding space \cite{Ospanov2025ScendiEmbeddings}.

This sophisticated approach is necessary because simple pixel-based or pairwise distance metrics fail when images are semantically redundant but visually distinct (e.g., two images of the same static scene taken seconds apart with minor lighting changes). 
By operating in the semantic feature space defined by CLIP, TCE ensures that diversity is penalized only when the selected images convey highly overlapping semantic content. We note, however, that spectral entropy in CLIP space serves as a proxy for human notions of visual diversity; the degree to which it captures all perceptually meaningful distinctions remains subject to further validation.

\subsection{Synthesizing the Composite Score}
The final step is to combine $S_{text}$, $S_{relevance}$, and $S_{diversity}$ into a single \texttt{MM-Eval} score. A simple average is inappropriate, as these sub-metrics differ in both scale and distribution ($S_{text}$ is a precision probability, $S_{rel}$ a Likert score, and $S_{div}$ a log-entropy value).

\subsubsection{The Regression Model}\label{sec:regression_model}
In order to learn the weights for all the \texttt{MM-Eval} components, we adopt the supervised learning approach pioneered by MMAE. We treat the aggregation as a regression problem where the goal is to predict the human judgment score.

\paragraph{Training Data:} We utilize the \textit{mLLM-Eval} Benchmark dataset. This dataset contains:
\begin{itemize}[noitemsep,leftmargin=*]
    \item Inputs: 142 distinct multimodal news articles.
    \item Outputs: Summaries generated by 9 different models (seq2seq, various MLLMs).
    \item Labels: $\sim1,500$ expert annotations providing ``Overall Quality'' scores ($y_{human}$).
\end{itemize}

\paragraph{Model:} We define the \texttt{MM-Eval} score as a linear combination of the normalized sub-scores:

\begin{equation}
\text{MM-Eval} = \beta_0 + \beta_1 S_{text} + \beta_2 S_{relevance} + \beta_3 S_{div}
\label{eq:mm_eval}
\end{equation}

We define the feature vector $X_i =$ for each sample $i$ in the benchmark. We then solve for the weights $\vec{\beta}$ that minimize the Mean Squared Error (MSE) against the human scores $y_{human}$:
\begin{equation}
    \hat{\beta} = \arg\min_{\beta} \sum_{i} \left( \beta^T X_i - y_{human}^{(i)} \right)^2
\end{equation}

Since the learned weights $\hat{\beta}$ reflect the preference structure of a specific evaluation context (here, news summarization), they may not directly transfer to domains where visual information plays a more central role. However, the aggregation model is deliberately shallow and low-dimensional, requiring only a small set of human preference judgments (e.g., ordinal ratings on tens to a few hundred examples) to recalibrate for a new domain. Crucially, the underlying component scorers, OpenFActScore, G-Eval, the MLLM judge, and TCE, remain applicable without modification, as they operate on the source document and system output alone and do not depend on domain-specific reference summaries.

\section{Analysis and Results}
\subsection{Implementation Details}\label{sec:implementation_details}
To obtain individual components scores for the three pillars of \texttt{MM-Eval}, we used Mistral-7B-Instruct for $S_{text}$, LLaVA-Mistral for $S_{diversity}$, and ViT for $S_{relevance}$. For learning the weights among components, we used Ridge Regression (with L2 regularization) and learned the weights in two stages, first stage for the different components involved in $S_{text}$ and then for the three components involved in Eq.~\eqref{eq:mm_eval} in stage 2. 
All scoring components used deterministic decoding (temperature 0, sampling disabled) to ensure reproducibility; further technical details are listed in Section~\ref{app:technical_details}.\footnote{Code: https://github.com/abidmeeraj/MM-Eval}

\subsection{Statistical Analysis of Annotations}
To rigorously justify the structure and parametrization of \texttt{MM-Eval}, it is necessary to analyze the underlying statistical properties of the human evaluation dataset. The descriptive statistics and visual distributions of the human scores reveal key patterns in evaluator consensus, model performance characteristics, and the points of difficulty in judging summary quality.

\begin{table}[!htbp]
\centering
\small
\begin{tabular}{lcc}
\hline
\textbf{Metric} & Mean ($\mu$) & Std. Dev. ($\sigma$) \\
\hline
Fluency & 4.156 & 0.631 \\
Coherence & 4.104 & 0.664 \\
Image-Set Quality & 3.871 & 0.709 \\
Overall Quality & 3.672 & 1.002 \\
Text--Images Relevance & 3.689 & 1.224 \\
Relevance & 3.574 & 1.295 \\
Consistency & 4.044 & 1.412 \\
\hline
\end{tabular}
\caption{Human evaluation descriptive statistics (N=1562).}
\label{tab:metric_stats}
\end{table}

\begin{figure*}[!tbp]
\vspace{-0.2cm}
    \centering
    \includegraphics[width=0.8\textwidth]{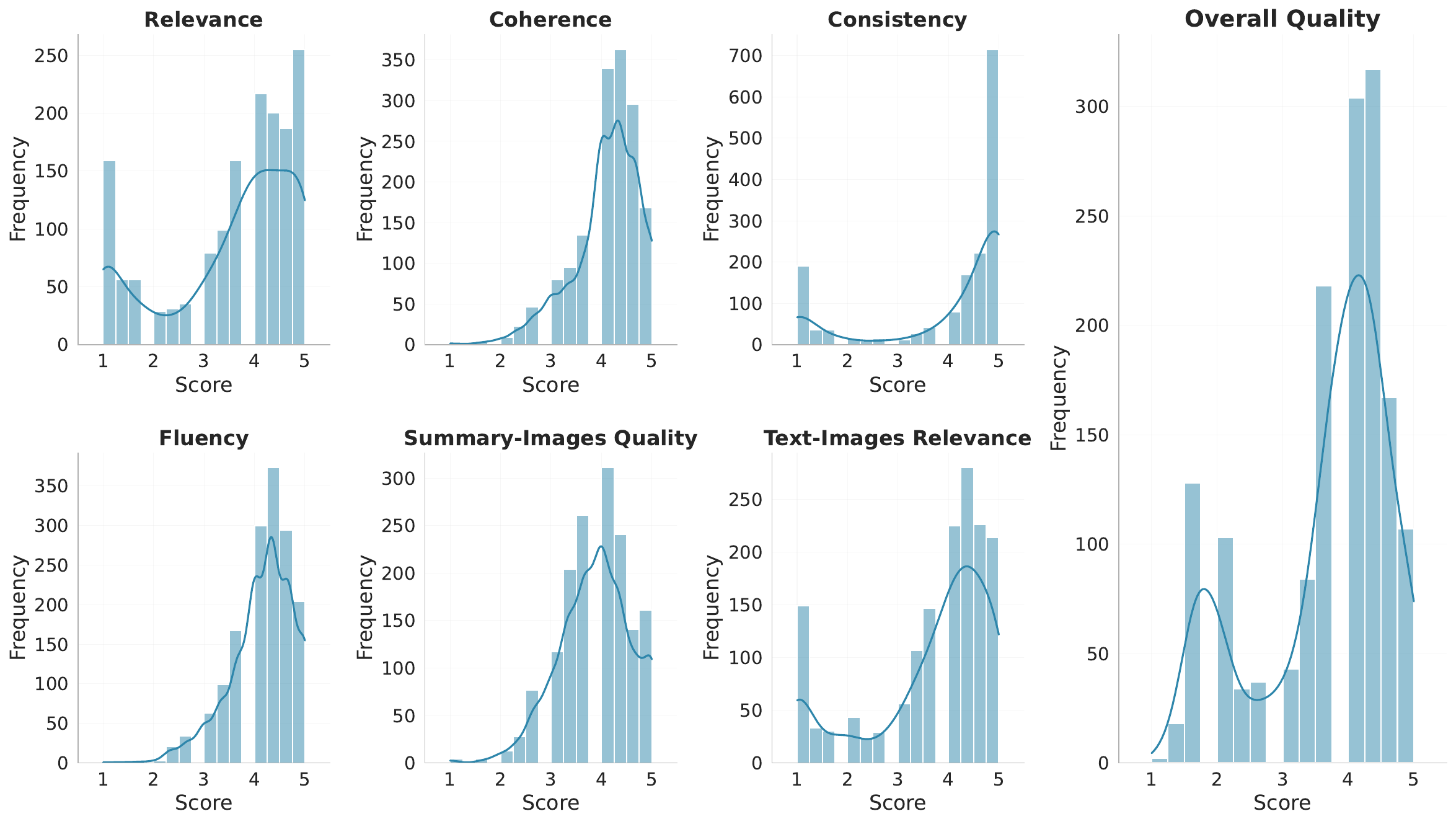}
    \caption{Human evaluation score distributions.}
    
    \label{fig:visual_distribution}
% \vspace{-0.4cm}
\end{figure*}

The human-annotations conducted by \textit{mLLM-Eval}, based on $N=1,562$ samples, provide a detailed look into how annotators perceived quality across seven dimensions and the same is described in Table~\ref{tab:metric_stats}. 

\paragraph{Linguistic Quality Consensus:}
Metrics concerning basic linguistic quality, specifically Fluency ($\sigma=0.631$) and Coherence ($\sigma=0.664$), as shown in Table~\ref{tab:metric_stats}, exhibit relatively low variance compared to other evaluated dimensions. The high mean scores ($\mu \approx 4.1$) combined with this limited dispersion indicate that system outputs are generally rated favorably along these dimensions, with comparatively little separation between models. As a result, fluency and coherence appear to function more as baseline quality indicators than as primary differentiators among systems in this evaluation setting.

\paragraph{The Intractability of Consistency and Relevance:} 
By comparison, dimensions related to information alignment and accuracy (Consistency ($\sigma=1.412$), Relevance ($\sigma=1.295$), and Cross-Modal Relevance ($\sigma=1.224$)), as shown in Table~\ref{tab:metric_stats}, exhibit substantially higher variance than linguistic quality metrics. This increased dispersion is consistent with prior findings that factual grounding and content selection are more challenging to assess reliably in summarization tasks \cite{Yuan2024EvaluateLLM}. In particular, the relatively large standard deviation observed for Consistency indicates greater variability in the assigned scores, which the regression model accounts for by assigning increased importance to $S_{fact}$ during aggregation.

\subsection{Interpretation of Score Distributions}

The distribution plots, given in Figure~\ref{fig:visual_distribution}, offer critical evidence explaining the regression model's learned hierarchy. While Fluency and Coherence exhibit distributions heavily concentrated toward the high end (suggesting ceiling effects), the distributions for Consistency, Relevance, and Overall Quality are widely spread or exhibit complex structures.

\paragraph{Bimodal Distribution of Factual Consistency}
The distribution plot for Consistency displays a pronounced bi-modal pattern, characterized by significant peaks at both the low end (scores 1-2) and the high end (scores 4-5). This distribution suggests two fundamentally different categories of model output: summaries that are factually sound (high scores) and summaries that contain significant, detectable hallucinations (low scores).

This bi-modal structure empirically supports the hypothesis of hallucination as a deal-breaker behavior in human evaluation. When a factual error is detected, the quality score plunges, regardless of how well the summary performs on secondary criteria like fluency or image quality. Conversely, outputs that maintain high factual integrity receive high scores. This clear separation confirms that Factual Consistency acts as a necessary gate function for overall quality. 

\begin{figure*}[!tbp]
    \centering
    \includegraphics[width=\textwidth]{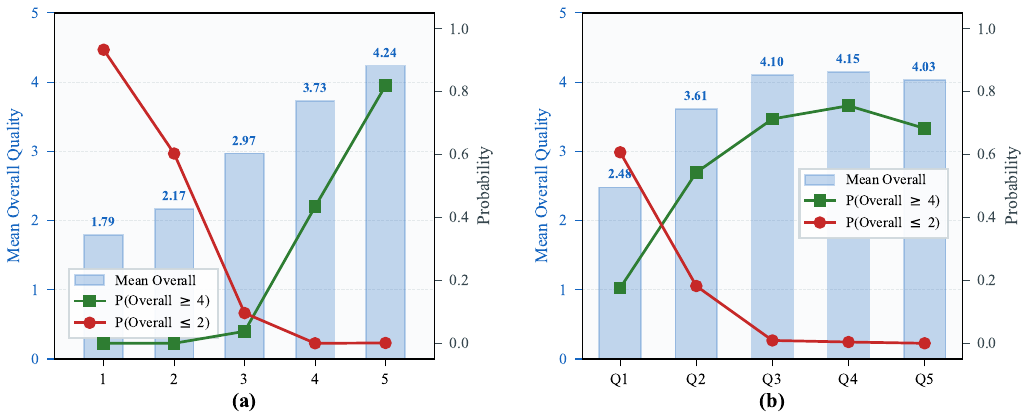}
    \caption{The gatekeeper effect of factual consistency on perceived overall quality. (a)~Human consistency bins versus overall quality ratings. (b)~Automatic factual consistency quintiles versus overall quality. In both panels, bars indicate
    mean overall quality (left axis). The green line tracks the probability of receiving a high overall rating ($\geq$4), while the red line tracks the probability of receiving a low overall rating ($\leq$2) (right axis). The
    crossover pattern illustrates that once factual consistency falls below a threshold, the likelihood of a poor overall score rises sharply regardless of performance on other dimensions.}
    \label{fig:gatekeeper_effect}
\end{figure*}

This gatekeeping effect is further illustrated in Figure~\ref{fig:gatekeeper_effect}, where both human consistency bins (panel~a) and automatic factual consistency quintiles (panel~b) exhibit a sharp crossover: as consistency increases, the probability of a high overall rating ($\geq$4) rises steeply while the probability of a low rating ($\leq$2) drops to near zero, reinforcing factuality as a critical determinant of perceived overall quality.

The Ridge regression model, by assigning a higher coefficient to $S_{fact}$, learns to mimic this penalty mechanism, ensuring that a low $S_{fact}$ score, corresponding to the low cluster in the human distribution, translates into a low \texttt{MM-Eval} score, mirroring the human response to fabricated information. 
This non-linear, threshold-like human response motivates a weighted metric that sharply penalizes factual failures rather than smoothing them away via uniform averaging.

\subsection{The Hierarchy of Weights: Interpretation and Implications}
The application of Ridge regression on the human-annotated dataset yields a crucial structural understanding of how annotators prioritize different quality dimensions in multimodal summarization. The resulting weight hierarchy is the most actionable output of the \texttt{MM-Eval} analysis, providing direct guidance for system optimization.

\begin{figure*}[!tbp]
    \centering
    \includegraphics[width=\textwidth]{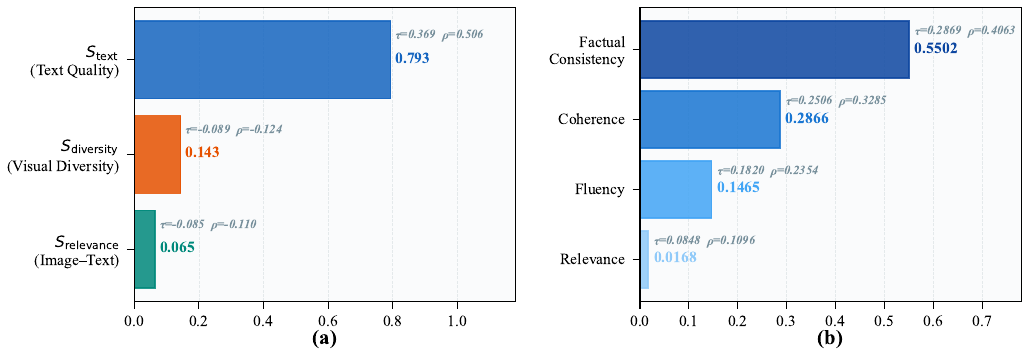}
    \caption{Learned weight hierarchy of \texttt{MM-Eval} components.
    (a)~Pillar-level weights with individual rank correlations ($\tau$, $\rho$) against human judgments.
    (b)~Intra-text component weights within $S_{\text{text}}$.}
    \label{fig:weight_hierarchy}
\end{figure*}

Figure~\ref{fig:weight_hierarchy}(a) visualizes the learned aggregation weights alongside per-component rank correlations, contrasting them against an equal-weighting baseline. Equal weighting nearly eliminates rank agreement with human judgments ($\tau=0.041$), indicating that annotators implicitly apply highly non-uniform priorities across dimensions; this motivates learning the aggregation weights rather than assuming uniform importance.

Table~\ref{tab:metric_summary} then reports the overall rank correlations and predictive fit statistics of the resulting model. We additionally verified that the learned weight ordering is stable under resampling: across repeated train/test splits, the dominance of $S_{\text{text}}$ and the centrality of factual consistency within text remain consistent, indicating the hierarchy is not an artifact of a single split.

We report rank correlations ($\tau$, $\rho$) because the downstream use of \texttt{MM-Eval} is comparative evaluation; absolute calibration is secondary to preserving human ranking preferences. Accordingly, the learned aggregation does not simply improve predictive performance, but more closely reflects the implicit trade-offs that humans apply when combining multiple criteria into a single overall judgment. Given the moderate magnitude of these correlations, \texttt{MM-Eval} is best suited for coarse-to-medium system comparisons; when systems are extremely close in quality, targeted human evaluation remains advisable.

\begin{table}[!tbp]
\centering
\small
\begin{tabular}{lccc}
\hline
\textbf{Metric} & \textbf{Value} & \textbf{95\% CI} \\
\hline
\textbf{Kendall's $\tau$} & 0.3744 & 0.374 [0.300, 0.444] \\
\textbf{Spearman's $\rho$} & 0.5139 & 0.514 [0.417, 0.597] \\
Pearson's $r$ & 0.6114 & -- \\
$R^{2}$ (Test Set) & 0.3719 & -- \\
RMSE (Test Set) & 0.8281 & -- \\
\hline
\end{tabular}
\caption{Overall correlation and error metrics with bootstrap confidence intervals.}
\label{tab:metric_summary}
\end{table}

\subsubsection{Learned Weights and The Textual Imperative}

The learned aggregation weights shown in Figure~\ref{fig:weight_hierarchy}(a) indicate a strong dominance of the textual component, with Text Quality ($S_{\text{text}}$) accounting for 79\% of the total aggregate weight. This is consistent with prior work emphasizing text as the core narrative medium in news summarization \cite{Zhu2020MultimodalReference}. Since Ridge Regression shrinks coefficients, lower values do not imply less importance and vice-versa \cite{James2023AnPython}. This hierarchy should not be interpreted as evidence that visual dimensions are unimportant in general; rather, it reflects the marginal contribution of each component to overall quality judgments in a domain where text carries the primary informational load.

Figure~\ref{fig:weight_hierarchy}(b) further decomposes $S_{\text{text}}$ into its constituent dimensions, revealing that factual consistency contributes the largest share of the text-level weight (0.55), followed by coherence (0.29), fluency (0.15), and relevance (0.02).

\subsubsection{Implications for Multimodal System Design}
The weight hierarchy derived from \texttt{MM-Eval} offers a data-driven view of how different quality dimensions contribute to human judgments in multimodal summarization. The learned weights suggest an implicit prioritization: ensuring factual grounding, followed by improvements in textual structure and coverage, and finally refinements to visual alignment and redundancy reduction, whose benefits appear conditional on strong textual quality.

\begin{table}[t]
\centering
\small
\begin{tabular}{p{0.55\linewidth} p{0.25\linewidth}}
\hline
\textbf{Component} & \textbf{Relative Contribution (\%)} \\
\hline
Factual Consistency & $\approx 43.5\%$ \\
Other Text Qualities & $\approx 35.5\%$ \\
Text Quality ($S_{\text{text}}$) & 79.0\% \\
Image-to-Text Relevance & $\approx 15.0\%$ \\
Visual Diversity (TCE) & $\approx 6.0\%$ \\
\hline
\end{tabular}
\caption{Learned component weights and their relative contribution to the $MM\text{-}Eval$ score.}
\label{tab:pillar_contributions}
\end{table}

Consistent with this ordering, Table~\ref{tab:pillar_contributions} shows that factual grounding accounts for the largest share of the aggregate score. Once textual quality is established, cross-modal alignment ($S_{relevance}$) contributes more substantially than aesthetic refinements such as visual diversity ($S_{diversity}$), which exhibit comparatively smaller marginal effects in this domain.

\begin{table}[!tbp]
\centering
\small
\begin{tabular}{lcc}
\hline
\textbf{Metric} & \textbf{Kendall's $\tau$} & \textbf{Spearman's $\rho$} \\
\hline
Factual Consistency & 0.2869 & 0.4063 \\
Coherence & 0.2506 & 0.3285 \\
Fluency & 0.1820 & 0.2354 \\
Relevance & 0.0382 & 0.0377 \\
Image-Text Relevance & \textbf{-0.0848} & \textbf{-0.1096} \\
Image-Set Diversity & \textbf{-0.0891} & \textbf{-0.1242} \\
\hline
\end{tabular}
\caption{Correlation strength of individual metrics.}
\label{tab:metric_correlations}
\end{table}

\begin{figure}[!tbp]
    \centering
    \includegraphics[width=\columnwidth]{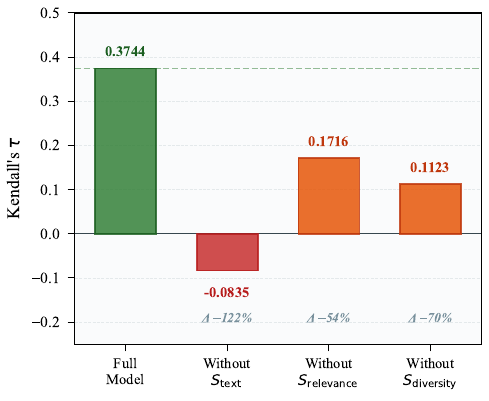}
    \caption{Ablation study showing each pillar's contribution to agreement with human rankings (Kendall's $\tau$). Removing $S_{\text{text}}$ causes agreement to drop below zero ($\Delta = -122\%$), while removing visual components also leads to substantial degradation.}
    \label{fig:ablation}
\end{figure}

Importantly, these priorities cannot be inferred from pairwise correlations alone.
Although Table~\ref{tab:metric_correlations} reports weak or negative associations for the visual proxies, the ablation results shown in Figure~\ref{fig:ablation} indicate that removing $S_{\text{relevance}}$ or $S_{\text{diversity}}$ leads to a marked drop in agreement with human rankings.
This suggests that visual components contribute in a conditional or interaction-dependent manner, rather than through monotonic effects. 

One plausible explanation is that, in text-dominant news summarization, visual redundancy or topical similarity does not consistently translate to perceived usefulness once the narrative is already conveyed by the text, and may in some cases align with less informative visual choices. More broadly, the discrepancy between marginal and joint effects is consistent with a conditional contribution pattern: visual components primarily refine quality distinctions when textual fidelity is already adequate, but may co-vary with lower-quality outputs when considered in isolation (e.g., systems that select topically similar but uninformative images may also produce weaker text). This conditional structure explains why removing these components degrades the joint model (Figure~\ref{fig:ablation}) despite their weak or negative marginal correlations (Table~\ref{tab:metric_correlations}),
and underscores the importance of evaluating multimodal quality jointly rather than dimension by dimension.

\subsection{Human Evaluation}\label{sec:human_eval}

A potential confound in interpreting the learned weight hierarchy is that the relatively low correlations observed for visual components may reflect limitations of the automatic proxies (TCE, MLLM judge) rather than genuine human indifference to these dimensions. To disentangle proxy noise from preference structure, we conducted a supplementary human evaluation on 200 randomly sampled benchmark articles. Three independent Amazon Mechanical Turk annotators rated each article, on a 5-point Likert scale, along four dimensions: text quality, image relevance, image diversity, and overall quality.

Table~\ref{tab:human_eval_complete} shows that annotators rated the visual dimensions favorably, with image relevance and diversity receiving scores comparable to text and overall quality. This indicates that the learned text-dominant hierarchy should not be interpreted as human indifference to visual quality. Rather, it reflects the \textit{marginal contribution} of each component to overall quality in this benchmark, where factual consistency acts as a gatekeeper. The agreement statistics further suggest that, although individual variation is expected, the annotations are sufficiently consistent for aggregate analysis. At the same time, the weaker automatic correlations for $S_{\text{relevance}}$ and $S_{\text{diversity}}$ may partly reflect proxy noise in TCE and the MLLM judge, motivating future work on stronger visual proxies.

\begin{table}[!tbp]
    \centering
    \small
    \setlength{\tabcolsep}{3.5pt}
    \begin{tabular}{lccc}
    \hline
    \textbf{Dimension} & \textbf{Mean} & \textbf{$\geq$4} & \textbf{Agreement} \\
    \hline
    Text & 3.90 (0.69) & 80.1 & 49.0 / 90.0 \\
    Image Relevance & 4.04 (0.80) & 76.8 & 44.3 / 84.0 \\
    Image Diversity & 3.89 (0.83) & 73.2 & 43.0 / 82.2 \\
    Overall & 4.00 (0.71) & 79.2 & 45.8 / 85.5 \\
    \hline
    \end{tabular}
    \caption{Human evaluation results. Mean reports mean score with standard deviation in parentheses; $\geq$4 reports the percentage of ratings at least 4; Agreement reports exact agreement / within-1 agreement across annotators.}
    \label{tab:human_eval_complete}
\end{table}

\section{Conclusion}
In this study, we proposed \texttt{MM-Eval}, a unified evaluation framework for multimodal summarization that integrates textual quality, cross-modal alignment, and visual diversity. Unlike prior approaches that rely on unimodal metrics or opaque model judgments, \texttt{MM-Eval} offers an interpretable, reference-weak alternative that aligns with human preferences through supervised aggregation.
Its modular design allows individual component scorers to be independently replaced or upgraded as stronger models become available.

Our analysis highlights a text-dominant hierarchy in the evaluation of news summaries, with factual consistency emerging as a critical determinant of overall quality.
A human evaluation confirms that annotators value visual relevance and diversity, indicating that the text-dominant weighting reflects the marginal contribution structure of the news domain rather than human indifference to visual dimensions. These findings validate the design of \texttt{MM-Eval} and offer practical guidance for system development: prioritizing factual grounding before optimizing visual components is likely to yield the largest gains in perceived quality within text-dominant settings.

\section*{Limitations}

While \texttt{MM-Eval} is validated on a text-dominant news summarization dataset, the learned weight hierarchy is shaped by domain-specific human evaluation behavior. In particular, the strong emphasis on factual consistency reflects the high variability and non-linear judgment patterns observed in this setting, where factual errors are heavily penalized relative to other dimensions. Furthermore, since most of the existing metrics rely on the reference-summaries as important component for evaluation, future work should assess the quality of reference summaries in existing datasets, in light of the new text-quality dimensions introduced.

The component scorers (OpenFActScore, G-Eval, the MLLM judge, and TCE) are reference-weak and domain-portable, as they operate on the source document and system output alone. When applying \texttt{MM-Eval} to a new domain, the learned aggregation weights from the news setting can serve as a reasonable default (zero-label transfer); if the target domain exhibits a substantially different preference structure, recalibrating only the aggregation model with a small set of human preference judgments (e.g., ordinal ratings on tens to a few hundred examples) provides a lightweight adaptation path. Future work should investigate the behavior of the framework in image-dominant domains (e.g., product reviews or technical documentation), where visual information plays a more central role. Re-calibrating the aggregation weights via Ridge regression in such settings would test whether \texttt{MM-Eval} can dynamically adapt its prioritization of modalities in response to different human evaluation contexts, further assessing its generality.

The automatic proxies used for visual relevance ($S_{\text{relevance}}$) and diversity ($S_{\text{diversity}}$) may not fully capture human perceptions of these dimensions. As shown in our human evaluation (Section~\ref{sec:human_eval}), annotators rate visual quality favorably, suggesting that the weaker automatic correlations for these components partly reflect proxy noise. Replacing TCE or the MLLM judge with stronger visual evaluation models may improve the fidelity of these pillars and shift the learned weight distribution.

Given the moderate magnitude of the observed rank correlations ($\tau=0.374$), \texttt{MM-Eval} is best suited for coarse-to-medium system comparisons. When systems are extremely close in quality, complementary targeted human evaluation remains advisable.

\texttt{MM-Eval} relies on multiple open-source LLMs and MLLMs, which increases computational cost relative to single-score lexical metrics. However, since evaluation is typically performed offline and far less frequently than model training or inference, this overhead is generally acceptable when reliable multimodal assessment is the goal. The modular design also permits practical trade-offs: users may substitute smaller or faster models for individual components, omit optional pillars (e.g., diversity) when resources are limited, or cache intermediate outputs such as atomic facts and CLIP embeddings to amortize repeated evaluations.

\section*{Ethical Considerations}
Our evaluation framework is trained on human preference data, which may encode subjective biases, such as an overemphasis on textual content. Additionally, reliance on LLMs introduces potential instability and opacity. We recommend using MM-Eval as a complement to, rather than a replacement for, human evaluation, particularly in high-stakes or sensitive domains.

\section*{Acknowledgments}
This research was undertaken with the assistance of resources from the National Computational Infrastructure (NCI Australia), an NCRIS enabled capability supported by the Australian Government.

\bibliography{references.bib}

\appendix

\section{Appendix}
\label{sec:appendix}

\subsection{Technical Details}\label{app:technical_details}
We experimented with different models for calculating the scores for the components involved in both the stages. The final models for each component were selected on the basis of best results on the given subset. The models are listed in Table~\ref{tab:model_components}.

\begin{table}[htbp]
\centering
\small
\begin{tabular}{l l}
\hline
\textbf{Component} & \textbf{Model(s)} \\
\hline
\multirow{4}{*}{Factual Consistency}
 & \textbf{Mistral-7B-Instruct-v0.1} \\
 & Mistral-7B-Instruct-v0.3 \\
 & Llama-2-7b-chat-hf \\
 & Llama-2-13b-chat-hf \\
\hline
\multirow{5}{*}{Rel, Coh, Flu}
 & \textbf{Mistral-7B-Instruct-v0.3} \\
 & Llama-2-7b-chat-hf \\
 & Llama-2-13b-chat-hf \\
 & Phi-3-mini-4k-instruct \\
 & Nous-Hermes-2-Mixtral-8x7B-DPO \\
\hline
\multirow{2}{*}{Relevance}
 & \textbf{Llava-v1.6-mistral-7b-hf} \\
 & Llava-onevision-qwen2-7b-ov-hf \\
\hline
\multirow{4}{*}{Diversity}
 & \textbf{ViT-B/32} \\
 & ViT-B/16 \\
 & ViT-L/14 \\
 & ViT-L/14@336px \\
\hline
\end{tabular}
\caption{Models used for each evaluation component. Final models used are highlighted in bold.}
\label{tab:model_components}
\end{table}

The obtained scores using these models were then normalized to a common range, the details to which are listed in Table~\ref{tab:normalization_bounds}. Its important to mention that we used TCE for the calculation of image-set diversity with the maximum eigen size of 20.

\begin{table}[htbp]
\centering
\small
\begin{tabular}{lll}
\hline
\textbf{Metric} & \textbf{Original} & \textbf{Normalized} \\
\hline
Factual Consistency & $[0,\,100]$ & $[0,\,1]$ \\
Other Textual Components & $[0,\,1]$ & $[0,\,1]$ \\
LLaVA Relevance & $[1,\,5]$ & $[0,\,1]$ \\
TCE Diversity & $[0,\,15]$ & $[0,\,1]$ \\
\hline
\end{tabular}
\caption{Normalization bounds applied to individual evaluation metrics.}
\label{tab:normalization_bounds}
\end{table}

\begin{figure*}[htbp]
    \centering
    \includegraphics[width=\textwidth]{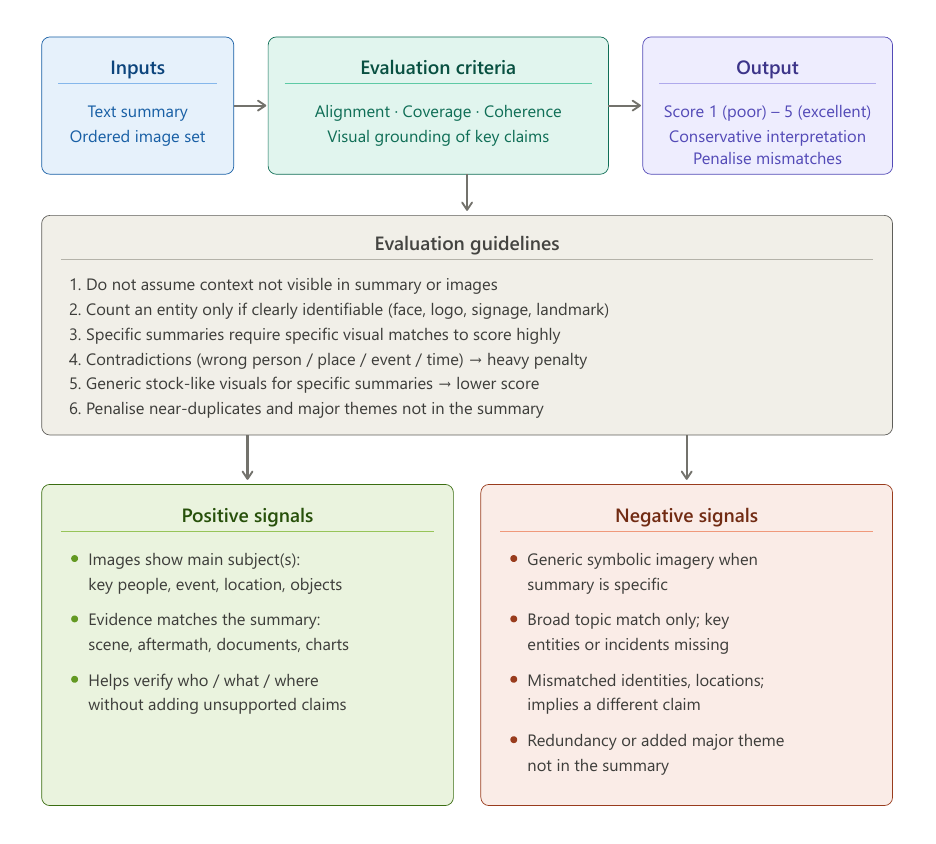}
    \caption{An Illustration of Image-Text Alignment scoring protocol.}
    \label{fig:alignment_prompt}
\end{figure*}

\paragraph{Prompts:} For $S_{text}$, we used the same prompts as G-Eval \cite{Liu2023G-EVAL:Alignment} except Factual Consistency, for which we relied on the default prompts in OpenFactScorer implementation. An illustration of prompt  used for $S_{relevance}$ is given as Figure~\ref{fig:alignment_prompt}.

\paragraph{Decoding Settings:} To ensure reproducibility and minimize run-to-run variance, all LLM- and MLLM-based scoring components (OpenFActScore, G-Eval, and the MLLM-as-a-judge for $S_{\text{relevance}}$) were configured with deterministic decoding: temperature was set to 0 and sampling was disabled. This departs from the original G-Eval protocol, which uses probability-weighted scoring over sampled tokens; we opted for deterministic outputs to prioritize consistency in an offline
benchmarking context. The models listed in Table~\ref{tab:model_components} were selected on the basis of best agreement with human judgments on the given subset; since
\texttt{MM-Eval} is modular, any component model can be substituted without altering the remainder of the pipeline.

As mentioned in Section~\ref{sec:implementation_details}, we learned the weights in two stages. For Stage 1 the $\alpha=1.0$, and for Stage 2 the $\alpha=0.1$ were selected via a 5-fold cross-validation. The dataset split was 80/20 which was stratified by the summarization system for the summaries in original dataset.\footnote{The singed coefficients for $S_{text}$, $S_{relevance}$, and $S_{diversity}$ were $2.7721$, $0.2256$, and $-0.4991$ respectively.}

\subsection{Stability Analysis}\label{app:stability_analysis}
This section provides detailed results about repeated-split end-to-end stability (S=50) where each repetition re-splits, re-learn Stage 1 + Stage 2, and evaluates on its own held-out split. Tables~\ref{tab:stability_rank}, \ref{tab:stability_pillars}, and \ref{tab:stability_intratext}, provide further evidence to the claims that the text-dominant hierarchy is not a single-split artifact: $w_{text}$ remains the largest weight across resamples and factual consistency remains the dominant text component.

\begin{table}[!tbp]
\centering
\small
\begin{tabular}{lrrrr}
\hline
\textbf{Metric} & \textbf{Mean} & \textbf{Std} & \textbf{2.5\%} & \textbf{97.5\%} \\
\hline
Kendall’s $\tau$ & 0.3788 & 0.0251 & 0.3382 & 0.4265 \\
Spearman’s $\rho$ & 0.5176 & 0.0312 & 0.4703 & 0.5732 \\
\hline
\end{tabular}
\caption{Stability of rank correlation across resampling runs.}
\label{tab:stability_rank}
\end{table}

\begin{table}[!tbp]
\centering
\small
\begin{tabular}{lrrrr}
\hline
\textbf{Weight} & \textbf{Mean} & \textbf{Std} & \textbf{2.5\%} & \textbf{97.5\%} \\
\hline
$w_{\text{text}}$ & 0.7572 & 0.0428 & 0.6748 & 0.8306 \\
$w_{\text{relevance}}$ & 0.0701 & 0.0215 & 0.0256 & 0.1091 \\
$w_{\text{diversity}}$ & 0.1727 & 0.0224 & 0.1431 & 0.2149 \\
\hline
\end{tabular}
\caption{Stability of learned pillar weights under resampling.}
\label{tab:stability_pillars}
\end{table}

\begin{table}[!tbp]
\centering
\small
\begin{tabular}{lrrrr}
\hline
\textbf{Text Weight} & \textbf{Mean} & \textbf{Std} & \textbf{2.5\%} & \textbf{97.5\%} \\
\hline
Factual Consistency & 0.5511 & 0.0085 & 0.5347 & 0.5644 \\
Coherence & 0.2870 & 0.0109 & 0.2693 & 0.3052 \\
Fluency & 0.1453 & 0.0125 & 0.1242 & 0.1664 \\
Relevance & 0.0166 & 0.0064 & 0.0025 & 0.0262 \\
\hline
\end{tabular}
\caption{Stability of intra-text weighting coefficients across resampling.}
\label{tab:stability_intratext}
\end{table}

\subsection{Detailed Numerical Results}\label{app:detailed_results}
Tables~\ref{tab:consistency_bins}--\ref{tab:ablation} provide the exact 
numerical values corresponding to Figures~\ref{fig:gatekeeper_effect}--\ref{fig:ablation}, in the main text.

\begin{table*}[!tbp]
\centering
% \small
\begin{tabular}{crrrrr}
\hline
\textbf{Consistency bin} & \textbf{n} & \textbf{Mean Overall} & \textbf{Median Overall} & $\mathbf{P(\text{Overall} \ge 4)}$ & $\mathbf{P(\text{Overall} \le 2)}$ \\
\hline
1 & 225 & 1.79 & 1.67 & 0.000 & 0.933 \\
2 & 58  & 2.17 & 2.00 & 0.000 & 0.603 \\
3 & 52  & 2.97 & 3.00 & 0.038 & 0.096 \\
4 & 290 & 3.73 & 3.67 & 0.434 & 0.000 \\
5 & 937 & 4.24 & 4.33 & 0.819 & 0.001 \\
\hline
\end{tabular}
\caption{Relationship between human consistency bins and overall quality ratings.}
\label{tab:consistency_bins}
\end{table*}

\begin{table*}[!tbp]
\centering
\small
\begin{tabular}{lrrrrr}
\hline
\textbf{Factual Consistency quintile} & \textbf{n} & \textbf{Mean Overall} & \textbf{Median Overall} & $\mathbf{P(\text{Overall} \ge 4)}$ & $\mathbf{P(\text{Overall} \le 2)}$ \\
\hline
Q1 & 313 & 2.48 & 2.00 & 0.176 & 0.607 \\
Q2 & 313 & 3.61 & 4.00 & 0.543 & 0.182 \\
Q3 & 345 & 4.10 & 4.00 & 0.713 & 0.009 \\
Q4 & 282 & 4.15 & 4.33 & 0.755 & 0.004 \\
Q5 & 309 & 4.03 & 4.00 & 0.683 & 0.000 \\
\hline
\end{tabular}
\caption{Relationship between Factual Consistency quintiles and overall quality ratings.}
\label{tab:fact_quintiles}
\end{table*}

\begin{table}[!tbp]
\centering
\small
\begin{tabular}{lccc}
\hline
\textbf{Component} & \textbf{Weight} & $\tau$ & $\rho$ \\
\hline
\textbf{Final MM-Eval} & -- & $ 0.374^{*}$ & $0.514^{*}$\\
$S_{\text{text}}$ (Text Quality) & 0.793 & $ 0.369 $ & $0.506$ \\
$S_{\text{relevance}}$ (Image--Text) & 0.065 & $ -0.085 $ & $-0.110$ \\
$S_{\text{diversity}}$ (Visual) & 0.143 & $-0.089 $ & -0.124 \\
Equal Weights Baseline & 0.333 & $0.041 $ & $0.058$\\
\hline
\end{tabular}
\caption{Correlation of model components with human judgments.}
\label{tab:mm_eval_components}
\end{table}

\begin{table}[!tbp]
\centering
\small
\begin{tabular}{lccc}
\hline
\textbf{Metric} & \textbf{Weight} & $\tau$ & $\rho$ \\
\hline
\textbf{Factual Consistency} & 0.5502 & 0.2869 & 0.4063\\
Coherence & 0.2866 & 0.2506 & 0.3285\\
Fluency & 0.1465 & 0.1820 & 0.2354\\
Relevance & 0.0168 & 0.0848 & 0.1096\\
\hline
\end{tabular}
\caption{Correlation of each text-quality component with human scores.}
\label{tab:metric_weights}
\end{table}

\begin{table}[!tbp]
\centering
\small
\begin{tabular}{lcc}
\hline
\textbf{Ablation} & \textbf{Kendall's $\tau$} & \textbf{$\Delta$ from Full} \\
\hline
Full Model & 0.3744 & -- \\
Without $S_{\text{text}}$ & -0.0835 & \textbf{-122\%} \\
Without $S_{\text{relevance}}$ & 0.1716 & -54\% \\
Without $S_{\text{diversity}}$ & 0.1123 & -70\% \\
\hline
\end{tabular}
\caption{Ablation study showing each pillar’s contribution.}
\label{tab:ablation}
\end{table}

\end{document}